\documentclass[runningheads]{llncs}

\usepackage[T1]{fontenc}
\usepackage{graphicx}
\usepackage{amsmath}
\usepackage{multirow}
\usepackage{booktabs}
\usepackage{url}
\usepackage{array}
\usepackage{tabularx}
\usepackage[section]{placeins}

\usepackage{xcolor}

\begin{document}

\title{Before Blaming the LLM, Diagnose the Workflow: A Role-Level Difficulty Profile for Translating Natural Language into Ontology-Grounded Statements}

\titlerunning{Before Blaming the LLM, Diagnose the Workflow}

\author{Liang Zhang\inst{1}\orcidID{0000-0003-4201-8535} \and
Beno\^{i}t Daireaux\inst{1}\orcidID{0000-0002-6936-0552} \and
Sergey Alyaev\inst{1}\orcidID{0000-0002-2105-2067}}
\authorrunning{L. Zhang et al.}
\institute{NORCE Norwegian Research Centre, Nyg\r{a}rdsgaten 112, 5008 Bergen, Norway
\url{https://www.norceresearch.no/}\\
\email{\{lizh,beda,saly\}@norceresearch.no}
}

\maketitle

\begin{abstract}
Natural-language-to-domain-specific-language (NL-to-DSL) semantic parsing systems increasingly translate informal user descriptions into structured artifacts such as Structured Query Language (SQL) queries, Application Programming Interface (API) calls, workflows, formal specifications, and ontology-grounded domain-specific language (DSL) statements. This paper focuses on ontology-grounded Assertion Box (ABox) DSL generation: the translation of natural language domain descriptions into assertion-level DSL statements grounded in a pre-existing ontology vocabulary. In this setting, syntactic validity is often easier to achieve than semantic adequacy. Failures are frequently attributed to weak generator models, but recurring sources of difficulty are also structural: insufficient input, weak domain grounding, output template and vocabulary gaps, weak validation, and non-local repair.

We propose a role-level difficulty profile for diagnosing difficulty in ontology-grounded NL-to-DSL semantic parsing at the workflow level.
The profile organizes diagnostic variables across five recurring workflow roles: Input, Generator, Output, Validator, and Repairer.
The seven variables are input sufficiency, context availability, grounding complexity, template fit, vocabulary coverage, validation strength, and error observability.
Practitioners construct the difficulty profile through qualitative judgment over available evidence.
This experience-informed, non-exhaustive difficulty profile is intended to support workflow diagnosis and guide redesign or human-in-the-loop intervention.

\end{abstract}

\section{Introduction}

\subsection*{Motivation}

Modern semantic parsing has expanded from classical logical-form translation to broader natural language-to-domain-specific-language (NL-to-DSL) systems. These systems map natural language into structured, executable, or verifiable artifacts such as Structured Query Language (SQL) queries, Application Programming Interface (API) calls, workflows, formal specifications, and ontology-grounded domain-specific language (DSL) statements. Text-to-SQL research illustrates the value of schemas and execution feedback, while grammar-constrained generation and structured-output methods show that syntactic and schema constraints can reduce malformed outputs~\cite{ref_spider2,ref_exec_guided_sql,ref_picard,ref_grammar_constrained,ref_structeval}.

However, generation does not imply reliable automation. In this paper, \emph{difficulty} refers to workflow-level structural friction that makes an NL-to-DSL task hard to automate. It is not the same as model accuracy, benchmark score, or abstract task complexity. In ontology-grounded Assertion Box (ABox) DSL generation, a DSL artifact may be syntactically valid yet semantically inadequate. A system may produce well-formed assertions that select the wrong ontology construct, force a concept into a nearby but inappropriate category, omit a required relation, or pass syntactic checks while failing expert review. Such failures are often attributed to weak generator models, but recurring bottlenecks are also role-level and structural.

Practitioners building ABox DSL generation systems need to know whether a task is suitable for automation, where failures arise in the workflow, whether the bottleneck lies in the generator or in the task structure, what redesign is likely to reduce friction, and where human expertise should enter the loop. Existing evaluations often emphasize exact match, syntax validity, execution success, or model accuracy. These measures are useful, but they are insufficient for diagnosing why a task is hard.

\subsection*{Related Work and Positioning}

Existing ways of characterizing NL-to-ontology difficulty are mostly partial. Traditional ontology learning distinguishes task layers such as term extraction, concept formation, taxonomy discovery, relation extraction, and axiom learning~\cite{ref_ontology_learning_from_text}. Recent benchmarks for large language models (LLMs), such as LLMs4OL, evaluate term typing, taxonomy discovery, and non-taxonomic relation extraction across domains~\cite{ref_llms4ol}. These works provide important task-level difficulty proxies, but they primarily address ontology learning or ontology generation, where the target is often Terminological Box (TBox) construction or extension. We instead focus on ABox-level DSL generation against an existing ontology vocabulary.

Semantic parsing and Text-to-X benchmarks provide a second line of evidence. Text-to-SQL, Text-to-API, structured-output generation, and formal-specification translation expose recurring issues such as schema grounding, structural complexity, ambiguity, constrained decoding, and execution or solver feedback~\cite{ref_spider2,ref_exec_guided_sql,ref_bfcl,ref_squab,ref_structeval,ref_picard,ref_grammar_constrained,ref_requirement_formalisation,ref_req2ltl}. These works show that difficulty is not only a property of the model; it also depends on the input, target language, feedback channel, and evaluation protocol. However, they usually characterize difficulty at the task, benchmark, or metric level rather than as a workflow diagnosis for practitioners.

Ontology evaluation and human-in-the-loop ontology engineering form a third relevant background. Competency questions, Shapes Constraint Language (SHACL)-style validation, expert review, and LLM-assisted ontology engineering all address parts of the validation and repair problem~\cite{ref_competency_questions,ref_xpshacl,ref_oe_assist,ref_ontology_llm}. Yet these mechanisms are rarely connected to upstream sources of friction such as insufficient input, missing context, grounding complexity, template mismatch, or vocabulary gaps. This paper therefore treats existing literature as a set of useful difficulty proxies and contributes a role-level difficulty profile that connects observed difficulty to actionable redesign decisions.

This is an experience-informed position paper that proposes a non-exhaustive role-level difficulty profile for practitioners.
Our position is that difficulty should be diagnosed at the workflow level rather than attributed primarily to the generator model.
The difficulty profile is motivated by the development of an ontology-grounded ABox DSL generation system and by recurring failure patterns reported in related NL-to-DSL literature.
The variable set may be revised as additional systems expose other concerns.
We do not claim that the difficulty profile predicts model accuracy or empirically generalizes across all domains.
Instead, we claim diagnostic utility: the difficulty profile helps practitioners make structural sources of difficulty explicit before investing in large-scale experiments, model tuning, or fully automated pipelines.

The same task label can be easy under one configuration and difficult under another. Difficulty depends on the input distribution, ontology and DSL design, available context, generator, validator, repair mechanism, and quality standard. The difficulty profile is therefore intended for feasibility assessment, early redesign, validation planning, and human-in-the-loop placement.

\subsection*{Contributions}
The paper makes three contributions:
\begin{enumerate}
 \item We identify five recurring roles in an ontology-grounded generation workflow: Input, Generator, Output, Validator, and Repairer, and argue that difficulty should be diagnosed at the workflow level, not the Generator level.
 \item We derive seven non-exhaustive diagnostic variables, each requiring different intervention: clarification, context enrichment, candidate ranking, template design, ontology extension, layered validation, and repair tracing.
 \item By constructing a difficulty profile for a diagnostic example from a drilling-domain system, we illustrate how it can turn a broadly stated generation failure into more specific redesign and human-in-the-loop decisions.
\end{enumerate}

\section{Typical Workflow and Recurring Roles}

\subsection{Common NL-to-DSL Task Families}

NL-to-DSL semantic parsing covers a broad family of tasks in which natural language is mapped to a structured target language. Common examples include Text-to-SQL, where user questions are translated into database queries~\cite{ref_spider2,ref_exec_guided_sql}; Text-to-API, where requests are mapped to tool or function calls~\cite{ref_bfcl}; Text-to-Workflow, where informal goals are mapped to multi-step process descriptions~\cite{ref_workteam}; and natural language to formal specification, where requirements are mapped to logical or solver-checkable representations~\cite{ref_requirement_formalisation,ref_req2ltl}.

These task families differ in target language, feedback channel, and validation strength. SQL queries can often be executed against a database; API calls can be checked against tool signatures and responses; workflow outputs may be partially executable; and formal specifications may be checked by solvers or model checkers. Ontology-grounded DSL tasks are different because their outputs often encode conceptual or assertion-level modeling choices whose correctness cannot always be reduced to execution success.

\subsection{Typical Workflow and Recurring Roles}

\begin{figure}[htbp]
\centering
\includegraphics[width=\textwidth]{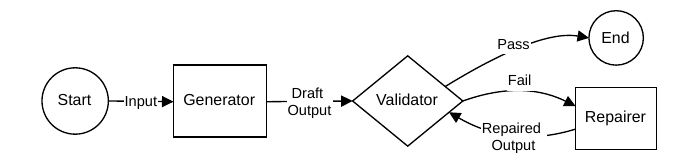}
\caption{Typical workflow and recurring roles in LLM-based ontology-grounded semantic parsing.}
\label{fig:typical_workflow}
\end{figure}

Figure~\ref{fig:typical_workflow} abstracts the recurring workflow.
The abstraction is implementation-neutral. A role may be implemented by a large language model (LLM) prompt, a retrieval module, a grammar-constrained decoder, a symbolic checker, a human reviewer, or a hybrid mechanism. The recurring roles are:
\begin{description}
    \item[Input] provides natural language intent and context. Typical failures include ambiguous, incomplete, implicit, or unsupported input.
    \item[Generator] grounds the input to target symbols and generates candidate DSL statements. Typical failures include wrong schema, tool, relation, or ontology-symbol selection, often because the generator cannot disambiguate among plausible target symbols.
    \item[Output] defines the DSL expression space and generated artifact. Typical failures include irregular output forms, ontology coverage gaps, awkward near-matches, and inadequate modeling choices.
    \item[Validator] checks whether the output is acceptable. Typical failures occur when syntax passes but meaning is wrong, or when feedback is weak, partial, or expert-only.
    \item[Repairer] revises, clarifies, regenerates, or escalates failures. Typical failures include non-local repair, repeated regeneration, and cases where ontology extension is needed.
\end{description}

\subsection{Focus: Ontology-Grounded ABox DSL Generation}

Ontologies are often described through the distinction between a Terminological Box (TBox) and an Assertion Box (ABox). The TBox defines the conceptual schema: classes, properties, taxonomies, constraints, and other ontology-level axioms. The ABox contains assertion-level statements about concrete individuals, facts, events, attributes, and relations under that schema.

In this paper, \emph{ontology-grounded semantic parsing} refers specifically to \emph{ontology-grounded ABox DSL generation}: mapping a natural language domain description, requirement, observation, event description, or user statement to DSL assertions that instantiate or connect ABox-level objects using a pre-existing ontology vocabulary.

At an abstract level, let \(f\) denote the end-to-end workflow, including its generation, validation, and repair roles. We write:
\[
y = f(x; O, G),
\]
where \(x\) is a natural language input, \(O\) is a pre-existing ontology, \(G\) is the DSL grammar, and \(y\) is the resulting set of ABox-level DSL assertions. This notation summarizes the workflow's input-output relation; it does not assume that \(f\) is a single model or a deterministic algorithm. The grammar specifies which declarations and assertions are syntactically well formed. In contrast, a \emph{template} or \emph{motif} is a recurring domain modeling pattern that may comprise several assertions. A candidate output should satisfy:
\[
\mathrm{valid}_G(y), \qquad \mathrm{vocab}(y) \subseteq O, \qquad \mathrm{adequate}_O(y, x).
\]
\(\mathrm{valid}_G(y)\) means that the output follows the DSL grammar. \(\mathrm{adequate}_O(y, x)\) means that, under ontology \(O\), \(y\) represents the intended meaning of \(x\) adequately. This last condition is a context-dependent acceptance judgment rather than an assumed fully automated decision procedure. The first two conditions are often easier to check than the third: an output can be syntactically valid and ontology-grounded while still using the wrong template, omitting required assertions, or choosing an inappropriate ontology construct.

This scope is narrower than ontology learning or ontology generation. Those tasks create or revise TBox-level vocabulary and axioms from text~\cite{ref_ontology_llm}. ABox DSL generation assumes that the ontology mostly exists and asks whether natural language can be mapped into assertion-level DSL statements grounded in that ontology. If the input contains an unsupported concept, the system may flag an ontology gap, but ontology extension is treated as a diagnostic outcome rather than as the primary generation target.

ABox DSL generation is difficult because syntax is only a weak success criterion. The output may be well formed but still use the wrong template, lack vocabulary coverage, omit required assertions, or require expert judgment to validate. These properties make it a useful focus for difficulty diagnosis.

\section{Role-Level Difficulty Profile}

The role-level difficulty profile is the main artifact of the paper. It organizes seven non-exhaustive diagnostic variables across five workflow roles. Each variable includes a diagnostic question, indicative evidence, and redesign options. Practitioners construct a case-specific profile through qualitative judgment of the available evidence. The resulting profile records where difficulty enters an NL-to-DSL workflow, what evidence supports the diagnosis, and what kind of redesign is likely to help. It is not a benchmark, a scoring metric, or a complete taxonomy of technical errors. Evidence may come from literature-reported failure modes, system logs, expert review comments, repeated wrong mappings, clarification questions, unsupported concepts, repair traces, escalation decisions, or case study observations.

The variables are diagnostic rather than orthogonal: a failure observed at one role may be caused or amplified by another role. They should therefore be read as prompts for diagnosis, not as statistically independent factors.
Each variable is organized as a diagnostic unit: a question to ask, evidence that the role is high-friction, and a redesign lever.

\begin{table}[htbp]
\centering
\caption{Overview of the role-level difficulty profile.}
\label{tab:profile_overview}
\small
\begin{tabularx}{\textwidth}{@{}p{3.3cm} p{1.5cm} p{4.5cm} X@{}}
\toprule
\textbf{Diagnostic variable} & \textbf{Role} & \textbf{High-friction evidence} & \textbf{Redesign lever} \\
\midrule
Input sufficiency & Input & Missing fields, implicit assumptions, ambiguous intent & Intent gate and clarification \\
Context availability & Generator & Missing terminology descriptions, examples, or modeling conventions & Knowledge base refinement \\
Grounding complexity & Generator & Many plausible nouns, verbs, paths, or templates & Candidate ranking strategy \\
Template fit & Output & Irregular or overlapping assertion patterns & Template or motif library \\
Vocabulary coverage & Output & Missing constructs or forced near-matches & Ontology extension mechanism \\
Validation strength & Validator & Semantic adequacy not deterministically checkable & Layered validation and human review \\
Error observability & Repairer & Failure cannot be localized or explained & Repair trace \\
\bottomrule
\end{tabularx}
\end{table}

The diagnostic units are as follows:
\begin{description}
    \item[Input sufficiency.] \textbf{Question:} Does the input contain enough explicit and interpretable information for mapping? \textbf{Evidence:} Fragmentary input, implicit assumptions, ambiguous intent, missing required fields, or several plausible mappings. Similar problems appear in requirements formalization and enterprise Text-to-SQL~\cite{ref_requirement_formalisation,ref_spider2,ref_practiq}; ambiguous and unanswerable semantic parsing tests further show that clean benchmark inputs can underestimate real input difficulty~\cite{ref_squab}. \textbf{Redesign:} Add an intent gate before generation. If required entities, relations, quantities, constraints, or disambiguating context are missing, ask a targeted clarification question or present candidate interpretations rather than forcing the input downstream.
    \item[Context availability.] \textbf{Question:} Does the generator have access to the domain and ontology context needed for generation? \textbf{Evidence:} Missing terminology descriptions, absent examples, poorly retrieved ontology fragments, or conventions buried in long context. Domain adaptation, long-context, and ontology-constrained extraction work show that model competence depends on accessible task-relevant context~\cite{ref_domain_adaptive_pretraining,ref_lost_middle,ref_relate}. \textbf{Redesign:} Refine the knowledge base by adding and maintaining descriptions for ontology terms and relations, curated examples, and modeling guidelines; then use ontology-aware retrieval to make the relevant parts available during generation.
    \item[Grounding complexity.] \textbf{Question:} Once relevant context is available, can the generator reliably ground the input to the right DSL symbols, templates, and argument structure? \textbf{Evidence:} Many plausible nouns, verbs, ontology paths, or templates; unstable candidate ranking; or multi-step symbol composition. Similar difficulty appears in Text-to-SQL schema linking, API/tool selection, workflow construction, and ontology-constrained relation grounding~\cite{ref_spider2,ref_bfcl,ref_workteam,ref_relate}. \textbf{Redesign:} Use a candidate ranking strategy to narrow the candidate space through ontology-aware retrieval, candidate generation and ranking, staged entity/relation/template selection, or even human confirmation of alternatives.
    \item[Template fit.] \textbf{Question:} Can valid outputs be represented by a small set of recurring templates, schemas, or grammar patterns? \textbf{Evidence:} Irregular composition, overlapping templates, or similar inputs that alternate between entity-centric, event-centric, relation-centric, or constraint-centric forms. Structured-output benchmarks and grammar-constrained decoding show the value and limits of explicit schemas and grammars~\cite{ref_structeval,ref_jsonschemabench,ref_grammar_constrained,ref_picard}. \textbf{Redesign:} Build an explicit template or motif library for recurring structural patterns, such as attribute assignment, relation assertion, event participation, constraint specification, state change, aggregation, or provenance attachment for generated assertions.
    \item[Vocabulary coverage.] \textbf{Question:} Does the ontology provide the classes, relations, attributes, and constraints needed to express the input within those templates? \textbf{Evidence:} Unsupported concepts, missing relations, forced near-matches, or locally defined terminology. Other NL-to-DSL settings face analogous output-side coverage constraints through schemas, signatures, action spaces, or formal languages~\cite{ref_spider2,ref_bfcl,ref_workteam,ref_req2ltl}; competency questions and ontology-generation studies expose similar coverage expectations~\cite{ref_competency_questions,ref_ontology_llm}. \textbf{Redesign:} Maintain an ontology extension mechanism so missing terms are detected, reviewed, and added when appropriate rather than silently replaced.
    \item[Validation strength.] \textbf{Question:} Can the system check whether the output is acceptable? \textbf{Evidence:} Syntax or type checks pass, but semantic adequacy remains uncertain; validation requires soft judgment or expert review. Execution-guided Text-to-SQL, tool-use evaluation, and formal-specification translation illustrate stronger feedback channels~\cite{ref_exec_guided_sql,ref_bfcl,ref_req2ltl}. In ontology settings, competency questions and ontology evaluation show why semantic adequacy often remains costly and expert-mediated~\cite{ref_competency_questions,ref_oe_assist}. \textbf{Redesign:} Use layered validation: deterministic syntax, type, domain/range, and template checks first, followed by LLM-assisted semantic review and human-in-the-loop confirmation for uncertain adequacy judgments.
    \item[Error observability.] \textbf{Question:} Can failures be localized and explained? \textbf{Evidence:} The artifact is rejected without revealing whether the cause is input insufficiency, missing context, wrong grounding, template mismatch, vocabulary gap, or weak validation. Repair-oriented LLM systems depend on actionable feedback~\cite{ref_self_refine,ref_reflexion}, and explainable SHACL work distinguishes machine-detectable violations from human-actionable explanations~\cite{ref_xpshacl}. \textbf{Redesign:} Preserve a repair trace that records the selected context, candidate symbols, templates, validation results, and human feedback. LLM-based judges or repairers may propose explanations and fixes, but semantic repairs should remain reviewable when they affect domain meaning or ontology evolution.
\end{description}

Context availability concerns whether the required domain knowledge is accessible to the generator, whereas grounding complexity concerns selecting among plausible ontology elements once that knowledge is available. The two may co-occur but suggest different interventions: context enrichment versus candidate ranking.

\section{Applying the Profile}

\subsection{Cross-Task Positioning}

We use the seven diagnostic variables to characterize difficulty in four better-known Text-to-X semantic parsing settings as comparison cases: Text-to-SQL, Text-to-API, Text-to-Workflow, and Text-to-formal-specification. The goal is not to rank tasks globally, but to show how ontology-grounded ABox DSL generation differs from better-known semantic parsing tasks in where friction tends to appear.
\begin{description}
    \item[Input sufficiency.] Text-to-API requests are often close to available tool signatures, and many Text-to-SQL benchmarks assume answerable questions over a given schema~\cite{ref_spider2,ref_bfcl}. Enterprise Text-to-SQL and ambiguous-query settings show that this assumption can break down~\cite{ref_practiq,ref_squab}. Ontology-grounded ABox DSL generation can present higher input friction when users provide short domain descriptions that omit required entities, relations, constraints, or modeling assumptions.
    \item[Context availability.] Text-to-SQL and Text-to-API commonly expose schemas, table names, function signatures, parameter descriptions, or examples as task context~\cite{ref_spider2,ref_bfcl}. ABox DSL generation can require richer context: ontology labels, relation meanings, domain conventions, modeling examples, and constraints that may not be present in the input or in the LLM's general knowledge.
    \item[Grounding complexity.] Text-to-SQL must ground language to tables, columns, joins, and query logic; Text-to-API must select tools and bind parameters; Text-to-Workflow must select and order workflow steps~\cite{ref_spider2,ref_bfcl,ref_workteam}. ABox DSL generation shares this grounding problem but may also require navigation among many plausible ontology classes, relations, paths, and templates, including assertion-carrier decisions about which individual should receive a relation.
    \item[Template fit.] Text-to-API has strong template fit when a tool signature directly defines the output shape, and Text-to-SQL benefits from a mature query language even when query structure is complex. Text-to-formal-specification may have difficult operator and temporal-structure choices, but the target formalism is explicit~\cite{ref_requirement_formalisation,ref_req2ltl}. ABox DSL generation can have weaker template fit when recurring modeling motifs are domain knowledge that has not yet been encoded into stable templates or grammar patterns. If these motifs are not defined, the same input may admit several plausible assertion structures.
    \item[Vocabulary coverage.] SQL schemas, API signatures, workflow action spaces, and formal languages make the available output vocabulary relatively explicit~\cite{ref_spider2,ref_bfcl,ref_workteam,ref_req2ltl}. ABox DSL generation depends on whether the ontology contains the needed classes, relations, units, quantities, and constraints. Because such ontologies are often domain-specific or case-specific, incomplete or under-defined vocabulary is common. If a required construct is absent, a syntactically valid output may still be a forced near-match rather than an adequate representation.
    \item[Validation strength.] SQL execution, tool-call traces, and formal checkers can provide strong feedback when outputs are executable or solver-checkable~\cite{ref_exec_guided_sql,ref_bfcl,ref_req2ltl}. ABox DSL generation can use syntax, type, path, SHACL-style, and competency-question checks, but \(\mathrm{adequate}_O(y, x)\) often remains expert-mediated~\cite{ref_competency_questions,ref_oe_assist,ref_xpshacl}.
    \item[Error observability.] Text-to-SQL execution errors, tool-call failures, and solver counterexamples can sometimes localize what went wrong. In ABox DSL generation, a rejected output may not reveal whether the failure came from insufficient input, missing context, wrong grounding, poor template fit, vocabulary gaps, or weak validation. This makes repair more dependent on repair traces, explanation, and human review.
\end{description}

Viewed through the seven variables, ontology-grounded ABox DSL generation differs from structured-output settings that benefit from stable schemas, explicit tool signatures, executable feedback, or solver checkers. In the system motivating this paper, sparse examples, incomplete knowledge resources, vocabulary gaps, and assertion structures that had not been stabilized as templates created friction across several variables. Two additional concerns were completeness and cross-assertion consistency: omitted assertions may matter even when every generated statement is valid, and one wrong type or relation can propagate through linked assertions. Repair may consequently require reinterpretation, expert escalation, or ontology extension rather than a local syntactic patch.

\subsection{Diagnostic Example}

\subsubsection{System setup.}
Consider an implemented ABox DSL generation system for drilling signals. The user provides a natural language description of a drilling signal, and the agent generates ontology-grounded DSL assertions in a DWIS-style configuration language, motivated by the Drilling and Wells Interoperability Standard (D-WIS) and the DDHub semantic interoperability framework~\cite{cayeux2022dwis,DWIS_DDHubSemanticInteroperability}. In the DDHub-derived vocabulary snapshot used by the prototype, the TBox contains roughly 2,250 class declarations, which we treat as noun-like ontology constructs, and roughly 200 object-property declarations, which we treat as verb-like relation constructs. These counts describe the vocabulary snapshot used by the prototype rather than an invariant property of D-WIS. The generated assertions may introduce signal individuals, measurable quantities, transformations, reference frames, physical locations, and links between them. The system retrieves ontology context, generates candidate assertions, validates syntax and type compatibility, and routes difficult cases to human review.

\subsubsection{A diagnostic case.}
The following diagnostic case compares an actual agent attempt with a reference configuration and illustrates why a short natural language request can still require a complex ontology-grounded ABox representation. It is not used as an evaluation instance. As a contrast, an easier request such as ``Create a moving average smoothed weight on bit signal located at the bottom of the string'' explicitly names the quantity, transformation, and location; it can be mapped to stable signal-to-quantity, signal-to-transformation, and signal-to-location templates. The following input is less explicit:

\begin{verbatim}
Give the semantics for retrieving the Bottom Hole Depth drilling signal.
\end{verbatim}

An actual agent attempt produced the following configuration:
\begin{verbatim}
DynamicDrillingSignal: bh_depth_signal
HoleBottomLocation: bh#01
DepthDrillingQuantity: bottom_hole_depth

bh_depth_signal IsOfMeasurableQuantity bottom_hole_depth
bh_depth_signal IsPhysicallyLocatedAt bh#01
\end{verbatim}

The reference configuration used in this diagnostic analysis is:
\begin{verbatim}
PhysicalData: BH_depth#01
DynamicDrillingSignal: BH_depth
OneDimensionalCurviLinearReferenceFrame: curvilinearAbscissaFrame#01
HoleBottomLocation: bh#01
DepthDrillingQuantity: bottom_hole_depth

BH_depth#01 HasDynamicValue BH_depth
BH_depth#01 IsOfMeasurableQuantity bottom_hole_depth
BH_depth#01 HasReferenceFrame curvilinearAbscissaFrame#01
BH_depth#01 IsPhysicallyLocatedAt bh#01
\end{verbatim}

The two configurations differ structurally. The actual attempt contains a dynamic drilling signal, a bottom-hole location, a depth quantity, and two assertions attached to the signal individual. Compared with this output, the reference configuration adds a \texttt{OneDimensionalCurviLinearReferenceFrame} individual and a \texttt{PhysicalData} individual, keeps a separate dynamic drilling signal individual, links the physical data individual to the signal through \texttt{HasDynamicValue}, and attaches measurable quantity, reference frame, and physical location assertions to the physical data individual. The reference is used for this diagnostic analysis rather than presented as a structure uniquely entailed by the short input alone.

\subsubsection{Constructing the difficulty profile.}
We diagnose this example along the seven variables as follows:
\begin{description}
    \item[Input sufficiency.] The input is concise but underspecified. It names ``Bottom Hole Depth'' but does not say whether the output should include the additional structures needed for this specific concept, such as a physical data object and a reference frame. In practice, users often assume that the agent can infer this implicit context from the domain phrase itself. The missing information is therefore not merely a missing slot; it is an implicit modeling assumption that the system must either infer or clarify.
    \item[Context availability.] The agent can only infer the intended structure if the relevant implicit modeling assumption has been supplied somewhere in the system context, such as examples, modeling guidelines, retrieved ontology documentation, or the LLM prompt. For a case-specific engineering team, it is difficult to prepare all such assumptions in advance. In this example, the agent must know that ``Bottom Hole Depth'' requires a reference frame.
    \item[Grounding complexity.] Even with relevant context, the mapping is not a direct lookup. The ontology contains many candidate classes and relations, so the system needs an effective strategy for multi-step grounding decisions. It must decide that \texttt{BH\_depth} is the dynamic signal, that \texttt{BH\_depth\#01} is the physical data object, that \texttt{HasDynamicValue} links them, and that reference-frame and physical-location assertions attach to the physical data object rather than to the dynamic signal.
    \item[Template fit.] The failed attempt illustrates weak template fit: because no stable template or motif fixes the reference structure for this case, the LLM produces an ad hoc signal-level-looking structure. The reference configuration uses a more complex assertion structure involving a physical data object, dynamic value link, measurable quantity, reference frame, and physical location. A case-specific team can prepare common templates, but it is difficult to anticipate all templates and template compositions in advance. The easier weight-on-bit example fits stable local templates; the bottom-hole-depth example requires a template or motif that makes this structure explicit.
    \item[Vocabulary coverage.] Since the necessary terms exist in the ontology, vocabulary coverage is not the dominant bottleneck in this particular case. However, the case would show greater friction in vocabulary coverage if any of these constructs were absent, because the generator would be forced into near-matches or ontology extension requests.
    \item[Validation strength.] A syntax or type checker can verify that relations and individuals are well formed, but it cannot fully establish \(\mathrm{adequate}_O(y, x)\). The actual configuration may satisfy simple syntactic checks, while the reference configuration assumes that \texttt{OneDimensionalCurviLinearReferenceFrame} and \texttt{PhysicalData} should also be represented. Because this modeling assumption is not explicit in the input, an LLM-based judge without access to the relevant context or templates may not identify the simpler output as inadequate.
    \item[Error observability.] Error observability is limited because validation itself cannot reliably determine semantic adequacy. If validation only reports that the output is semantically uncertain, or if an expert rejects it as incomplete, the repairer still may not know whether the cause is insufficient input, missing context, wrong grounding, weak template fit, or an ontology gap. Without a repair trace over retrieved context, candidate classes, selected templates, and grounding choices, repair becomes repeated regeneration rather than targeted correction.
\end{description}

\subsubsection{Redesign implications.}
The point of this example is not to show that a redesigned agent outperforms a baseline. It shows how practitioners can use the profile to translate a vague failure ``the generated semantics is not quite right'' into role-level design actions.

\textbf{Input.} The system should include an intent gate before generation. For example, in the above bottom-hole-depth request, the gate may check whether the input provides enough information for the structures expected by this concept, such as the needed reference frame and physical location. The general principle is that if the input does not provide enough information to select the intended modeling structure, the system should ask a targeted clarification question or present candidate interpretations rather than silently choosing one.

\textbf{Generator and Output.} The system should not rely on a stronger model alone to infer case-specific implicit modeling assumptions. The system builder should encode domain knowledge explicitly, including recurring modeling assumptions, terminology descriptions, modeling guidelines, and curated examples. The generator should apply a candidate ranking strategy, for example by using ontology-aware retrieval to narrow the candidate space. It should also expose a repair trace, such as why a physical data object was selected rather than a simple dynamic signal assertion. On the output side, the system should use a template or motif library to guide the LLM toward stable assertion structures, instead of letting it invent an output pattern for each request. For example, the library can encode when a signal can be represented by a simple signal-to-quantity/location/transformation structure and when a specific concept requires additional structures such as physical data, dynamic value, measurable quantity, reference frame, and location. If a needed noun, verb, or unit is absent, the system should surface a vocabulary-coverage issue and route it to an ontology extension mechanism rather than inventing a near-match.

\textbf{Validator and Repairer.} Deterministic validation can verify syntax, relation names, types, and template conformance, but it cannot fully establish \(\mathrm{adequate}_O(y, x)\). When the remaining uncertainty is semantic, the realistic redesign is not fully automatic repair but human-in-the-loop review. An LLM can prepare a review package for the expert by highlighting the selected template, retrieved context, grounding choices, missing expected assertions, possible alternatives, and any suspected vocabulary gap. The expert still decides whether the output is semantically adequate before it enters downstream workflows.

\section{Limitations}

The difficulty profile supports qualitative diagnosis rather than performance prediction.
It does not forecast system performance, compare models, or establish that one pipeline outperforms another. Because the difficulty profile is grounded in practitioner experience and literature synthesis, its claims should be read as analytic and design-oriented rather than statistically generalizable. The seven diagnostic variables are non-exhaustive and should not be treated as independent factors; for example, low context availability can amplify grounding complexity, and weak template fit can interact with vocabulary coverage. Role-level judgments may also vary across evaluators.

The difficulty profile also does not replace task-specific quantitative evaluation. A production system still needs task-appropriate validation, expert review, and acceptance criteria. The difficulty profile has not yet been rigorously validated through quantitative studies. More case studies are needed to test whether independent practitioners construct similar difficulty profiles, whether using the profile improves redesign decisions, and how well it transfers across domains. ABox DSL systems may vary substantially in ontology maturity, template regularity, validation strength, and domain risk. The intended scope assumes that the ontology is at least partially available, while the fully open-ended ontology construction from scratch is outside the main scope. As LLM-based modeling systems and ontology engineering tools evolve, future work may refine the variable set or add new diagnostic variables when recurring bottlenecks are observed across additional cases.

\section{Conclusion}

This paper examined how LLM-based systems convert a natural-language description into formal, machine-readable statements in a domain-specific language, using only concepts and relations already defined in an existing ontology.
In the drilling example, such statements specify which signal is meant, what quantity it measures, and where it is located.
We observe that LLMs' ability to generate domain-specific statements does not imply reliable automation. Generation errors may stem from workflow-level bottlenecks, such as insufficient input, missing context, unstable output structures, or weak validation, rather than LLM capability alone.

Using the example from drilling, we showed that splitting a typical workflow into five roles can clarify the needed improvements. These conceptual roles are Input, Generator, Output, Validator, and Repairer.
We assessed them against seven non-exhaustive diagnostic variables: input sufficiency, context availability, grounding complexity, template fit, vocabulary coverage, validation strength, and error observability.
The resulting role-level difficulty profile is intended as a practical aid for locating structural friction and linking it to redesign options such as clarification gates, knowledge base refinement, candidate ranking strategies, motif libraries, ontology extension mechanisms, layered validation, and human review.
The main takeaway is simple: before treating a domain-specific-language generation failure as a generator-LLM failure, practitioners should first diagnose where the workflow itself may make the task difficult.

\section*{Acknowledgments}


This work is part of the Center for Research-based Innovation DigiWells: Digital Well Center for Value Creation, Competitiveness and Minimum Environmental Footprint (NFR SFI project no. 309589, https://DigiWells.no).
The center is a cooperation of NORCE Norwegian Research Centre, the University of Stavanger, the Norwegian University of Science and Technology (NTNU), and the University of Bergen. 
It is funded by Aker BP, ConocoPhillips, Equinor, Harbour Energy, Petrobras, TotalEnergies, Vår Energi, and the Research Council of Norway.

\bibliographystyle{splncs04}
\bibliography{references}

\end{document}